\pdfoutput=1

\RequirePackage{snapshot}
\documentclass[11pt]{article}

\usepackage{EMNLP2022}
\usepackage{hyperref}
\usepackage[capitalise]{cleveref}
\usepackage{times}
\usepackage{latexsym}
\usepackage{algorithm}
\usepackage{algpseudocode}
\usepackage{graphicx}
\usepackage[T1]{fontenc}

\usepackage[utf8]{inputenc}

\usepackage{microtype}

\usepackage{inconsolata}
\usepackage{booktabs}
\usepackage{multirow}
\usepackage{subcaption}

\title{Improving Stability of Fine-Tuning Pretrained Language Models \\ via Component-Wise Gradient Norm Clipping}

\author{Chenghao Yang$^1$, Xuezhe Ma$^2$ \\
    $^1$University of Chicago \\
    $^2$University of Southern California \\
    \texttt{yangalan1996@gmail.com}, \texttt{xuezhema@isi.edu}
}

\begin{document}
\maketitle
\begin{abstract}
Fine-tuning over large pretrained language models (PLMs) has established many state-of-the-art results. 
Despite its superior performance, such fine-tuning can be unstable, resulting in significant variance in performance and potential risks for practical applications. 
Previous works have attributed such instability to the catastrophic forgetting problem in the top layers of PLMs, which indicates iteratively fine-tuning layers in top-down manner is a promising solution. 
In this paper, we first point out that this method does not always work out due to different convergence speeds of different layers/modules. 
Inspired by this observation, we propose a simple component-wise gradient norm clipping method to adjust the convergence speed for different components. 
Experiment results demonstrate that our method achieves consistent improvements in terms of  generalization performance, convergence speed and training stability. The codebase can be found at  \url{https://github.com/yangalan123/FineTuningStability}.
\end{abstract}

\section{Introduction}

Fine-tuning over large pretrained language models (PLMs), which achieved remarkable performance over various benchmarks, has become the de facto paradigm for several current natural language processing (NLP) systems.
However, fine-tuning can be unstable in terms of significant variance in metrics, resulting in even worse-than-random failed models~\citep{devlin-etal-2019-bert, lee2019mixout, dodge2020fine, mosbach2020stability}. 

Catastrophic forgetting~\citep{kirkpatrick2017overcoming} during fine-tuning of PLMs is one common explanation for this instability~\citep{lee2019mixout}, i.e., PLMs may lose their rich  domain-agnostic knowledge acquired by language model pretraining in the process of fine-tuning. 
Through layer-replacement experiments between pretrained models and fine-tuned models,  \citet{mosbach2020stability} further connected the catastrophic forgetting problem to the optimization problem on top layers. 

These findings give rise to a straightforward way to enhance the fine-tuning stability: how about fine-tuning the model from top to bottom to reduce the parameter changes and hence mitigate the catastrophic forgetting problem? 
This is reminiscent of the gradual unfreezing~\citep{howard2018universal}, which does layer-wise top-down fine-tuning and unfreezing new layers only when the layers above have been fine-tuned. 
Therefore, the newly unfrozen layer would only be tuned for a slightly easier optimization problem at each iteration, leading to  much fewer changes to the parameters. 
\begin{figure}[tbp!]
\centering
\includegraphics[width=\columnwidth]{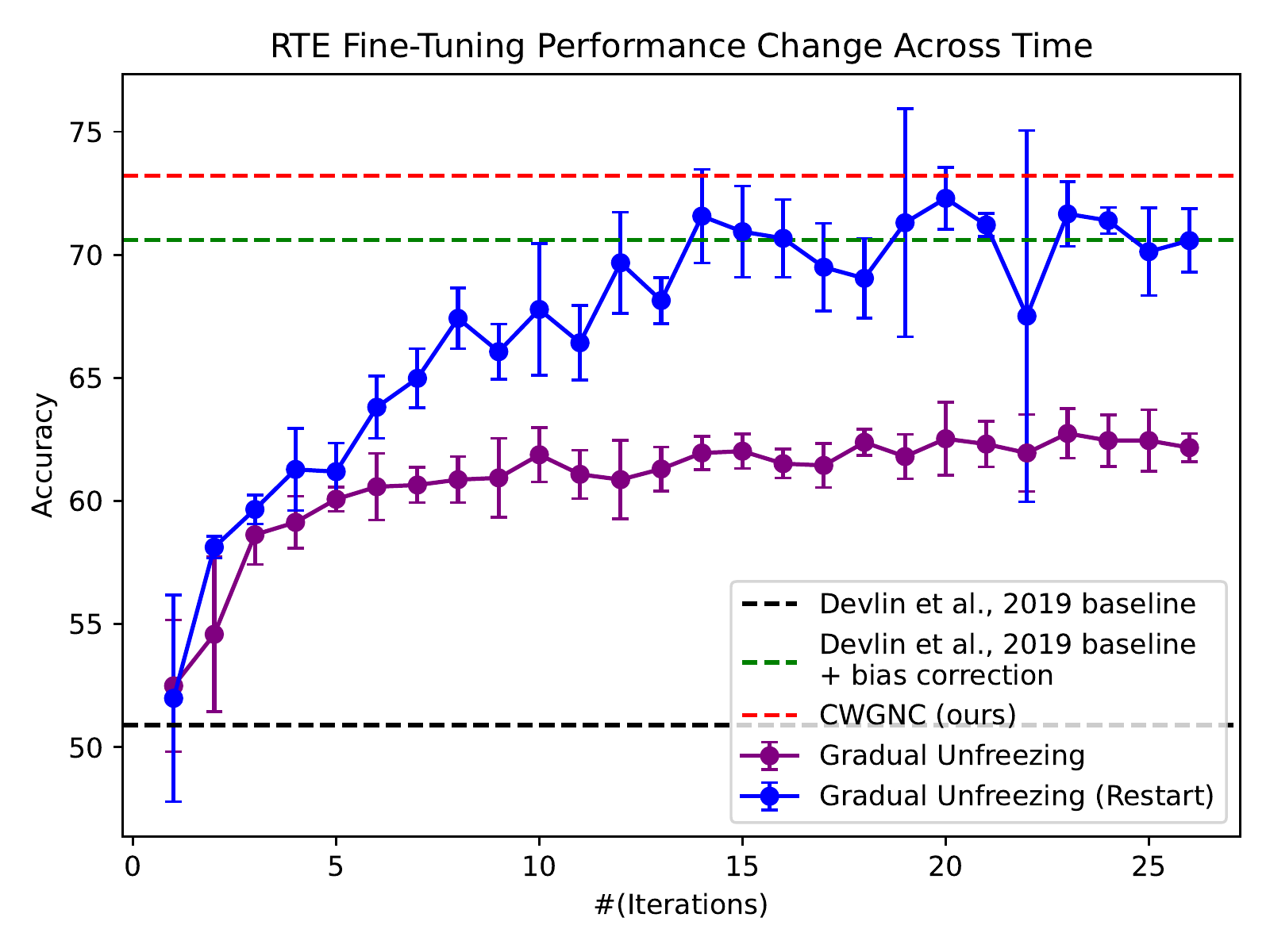}
\caption{Fine-tuning performance over time on RTE datasets. Here we fine-tune over BERT-large-uncased model \citep{devlin-etal-2019-bert}. In each iteration, we train the model for $3$ epochs and the errorbar is plotted based on $5$ different runs. }
\label{fig: perf_over_time}
\vspace{-3mm}
\end{figure}

However, based on a comprehensive case study of the gradual unfreezing method, we obtained empirical observations beyond our expectations (\S\ref{sec:case-study}). 
Our analysis further reveals a possible reason: \emph{different components (e.g., feed-forward networks at different layers, fully connected matrices and biases at output layer) converge at varying speeds.}
Thus, components in upper layers, which have converged to local optima, cannot easily be fine-tuned with newly unfrozen parameters. 

\begin{figure*}[t]
\centering
\begin{subfigure}[b]{0.48\textwidth}
\centering
\includegraphics[width=\columnwidth]{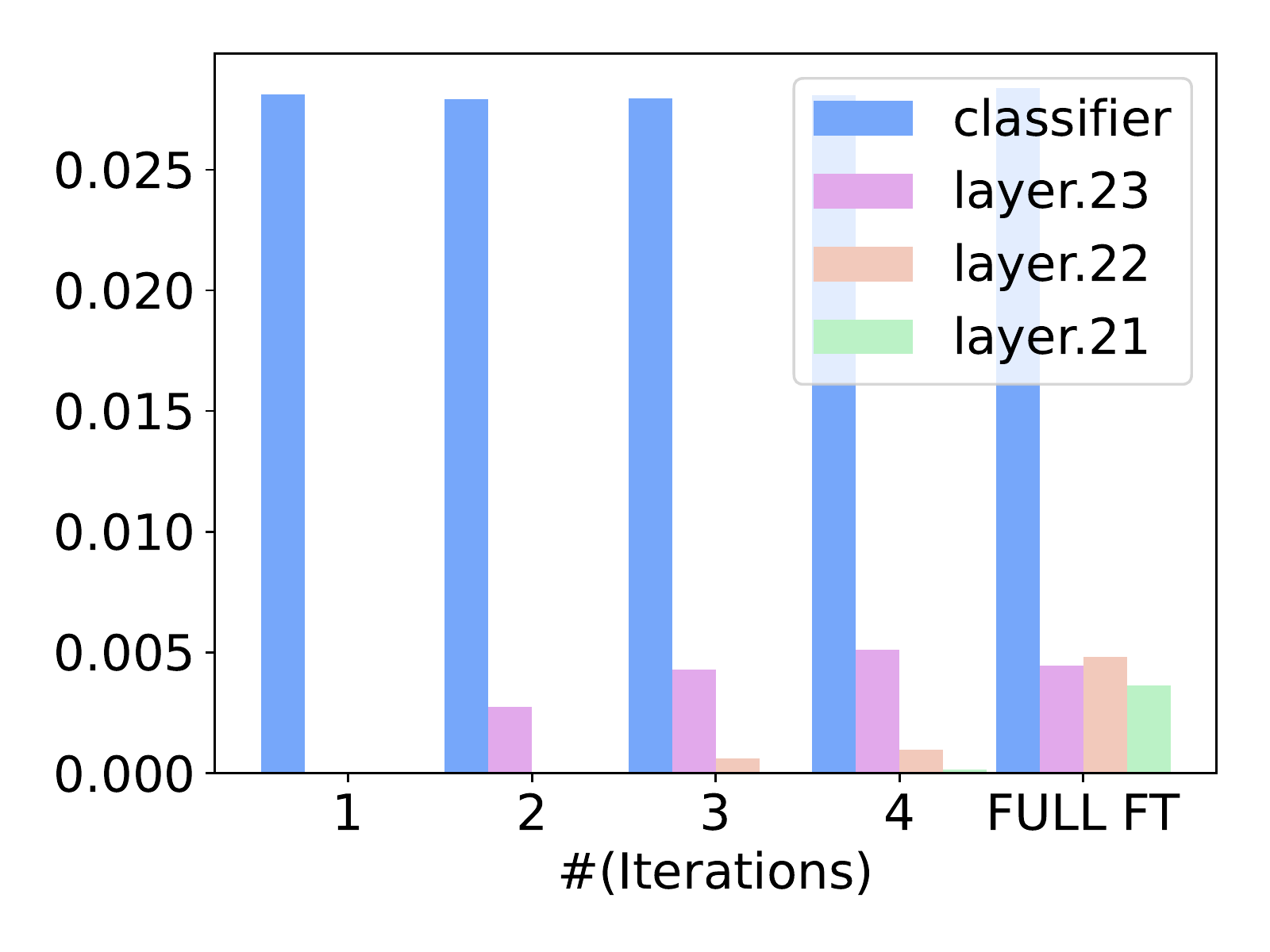}
\caption{ {Compared with original BERT model} }
\label{fig: raw_norm_over_time}
\end{subfigure}
\begin{subfigure}[b]{0.48\textwidth}
\centering
\includegraphics[width=\columnwidth]{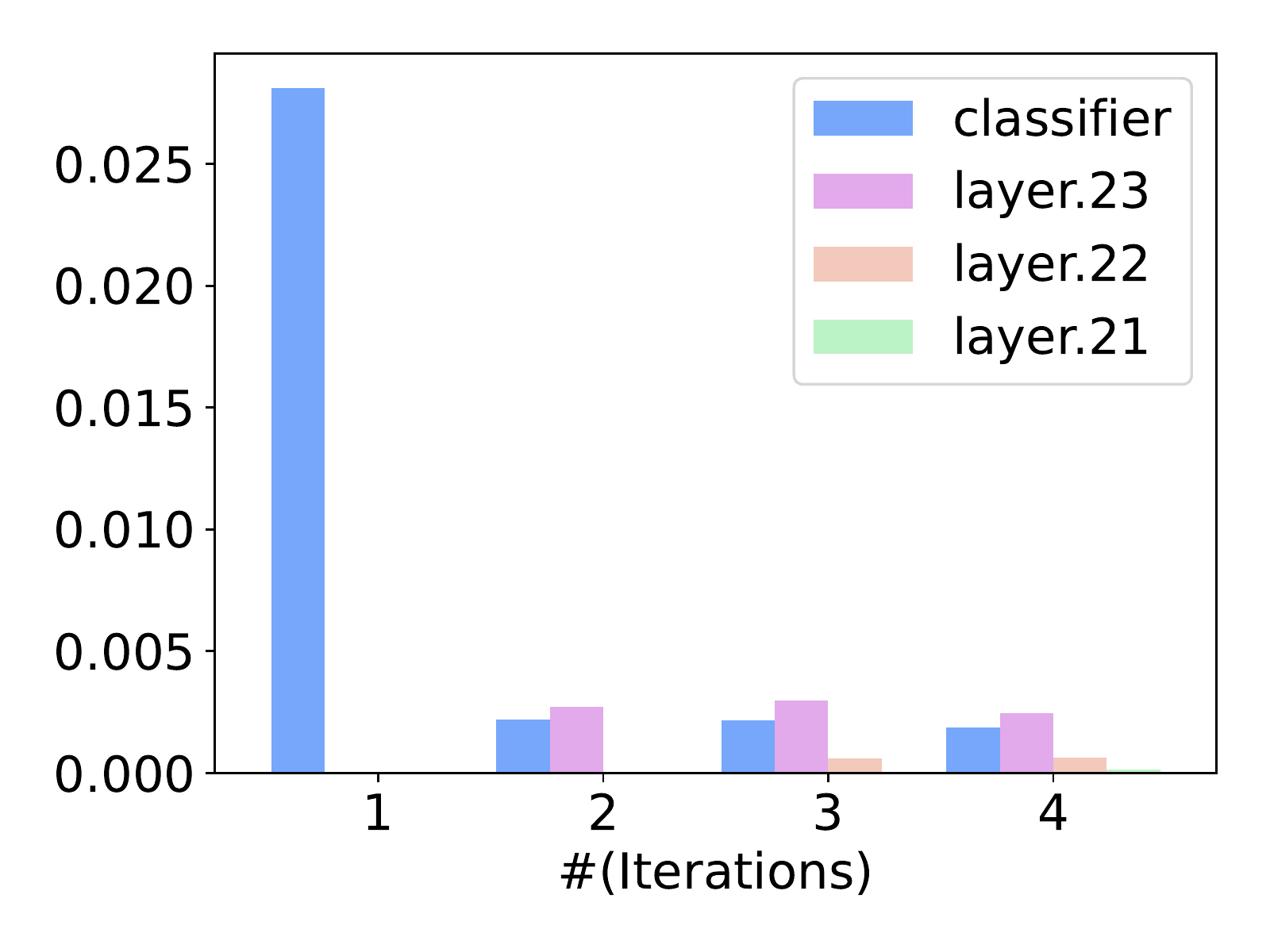}
\caption{{Compared with previous iteration} }
\label{fig: inc_norm_over_time}
\end{subfigure}
\caption{Parameter update at each iteration for GU. Updates that are too small cannot be seen in this figure (e.g., for the $22$-th layer, or ``layer.21'' in the figure update is too small to plot out in \cref{fig: inc_norm_over_time}). We only plot the first $4$ iterations as we observe that the performance is almost stable by the fourth iteration. As different layers can have components with different dimensionalities, we show the component-wise maximum rooted mean squared difference for each layer.}
\end{figure*}

Based on this observation, we propose a simple component-wise gradient clipping method to stabilize the fine-tuning process (\S\ref{sec:method}).
This method achieves significant empirical improvements of fine-tuning stability in terms of the variance and the failed run percentage over three tasks (\S\ref{sec:experiment}). 
In summary we make the following contributions:
\begin{enumerate}
    \item We find that component-wise convergence speed divergences is the key challenge in fine-tuning stability, based on the case study of the gradual unfreezing method. 
    \item Based on our observation, we propose a new simple component-wise gradient clipping method to help stabilize fine-tuning, which achieves empirical improvements of fine-tuning stability over previous methods.
\end{enumerate}

\section{A Bitter Case Study: Layer-wise Gradual Unfreezing}
\label{sec:case-study}

\citet{mosbach2020stability} attributed the instability problem in fine-tuning process to the catastrophic forgetting in the top layers, through a layer replacement experiment between pretrained and fine-tuned models. 
Following this empirical observation, the instability problem might be mitigated if we can minimize the edits to the pretrained model parameters, especially in top layers. 
This inspires us to mitigate the instability problem via gradual unfreezing (GU, \citet{howard2018universal}).

Specifically, suppose we are working with a model $M$ with $L$ layers parameterized by $\{\theta^{(i)}, 1\leq i \leq L\}$. 
GU tunes $M$ for $L$ iterations.
At $k$-th ($k$ start from $0$) iteration, we only tune a subset of parameters $R^{(k)} = \{\theta^{(i)}, L-k \leq i \leq L\}$, where $\theta^{(i)} (L-k+1 \leq i \leq L)$ is also tuned in $k-1$-th iteration. 
In each iteration, we tune the parameter for $E$ epochs, where $E$ is large enough for convergence.\footnote{We select $E$ based on preliminary experiments.} 
Detailed algorithm is shown in \cref{alg:gradual_unfreezing} at \cref{sec: algo_GU}.

\begin{figure*}[t]
\centering
\begin{subfigure}[b]{0.45\textwidth}
\centering
\includegraphics[width=\columnwidth]{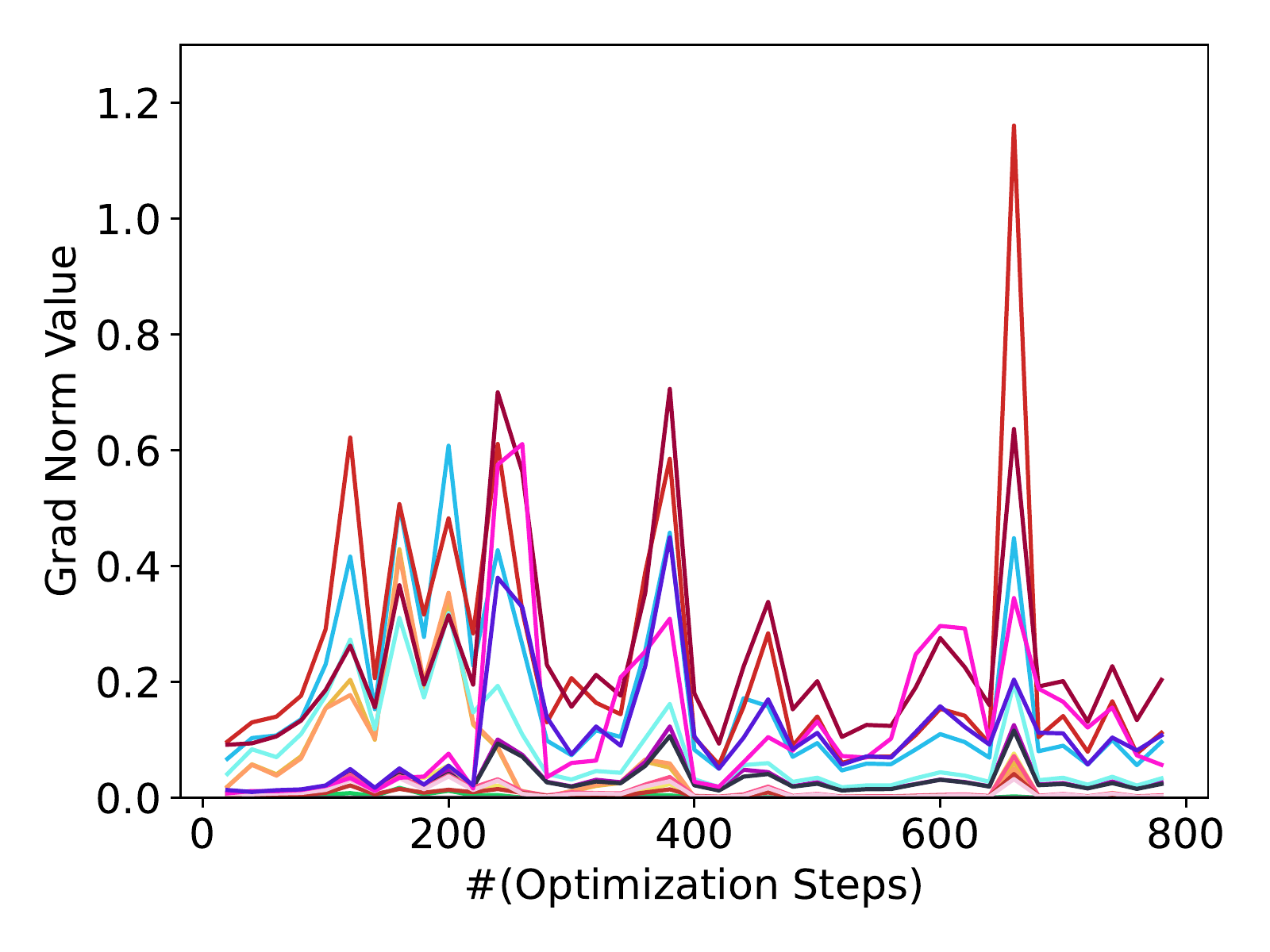}
\caption{ {Failed Run} }
\label{fig: failed_run_grad_norm}
\end{subfigure}
\begin{subfigure}[b]{0.45\textwidth}
\centering
\includegraphics[width=\columnwidth]{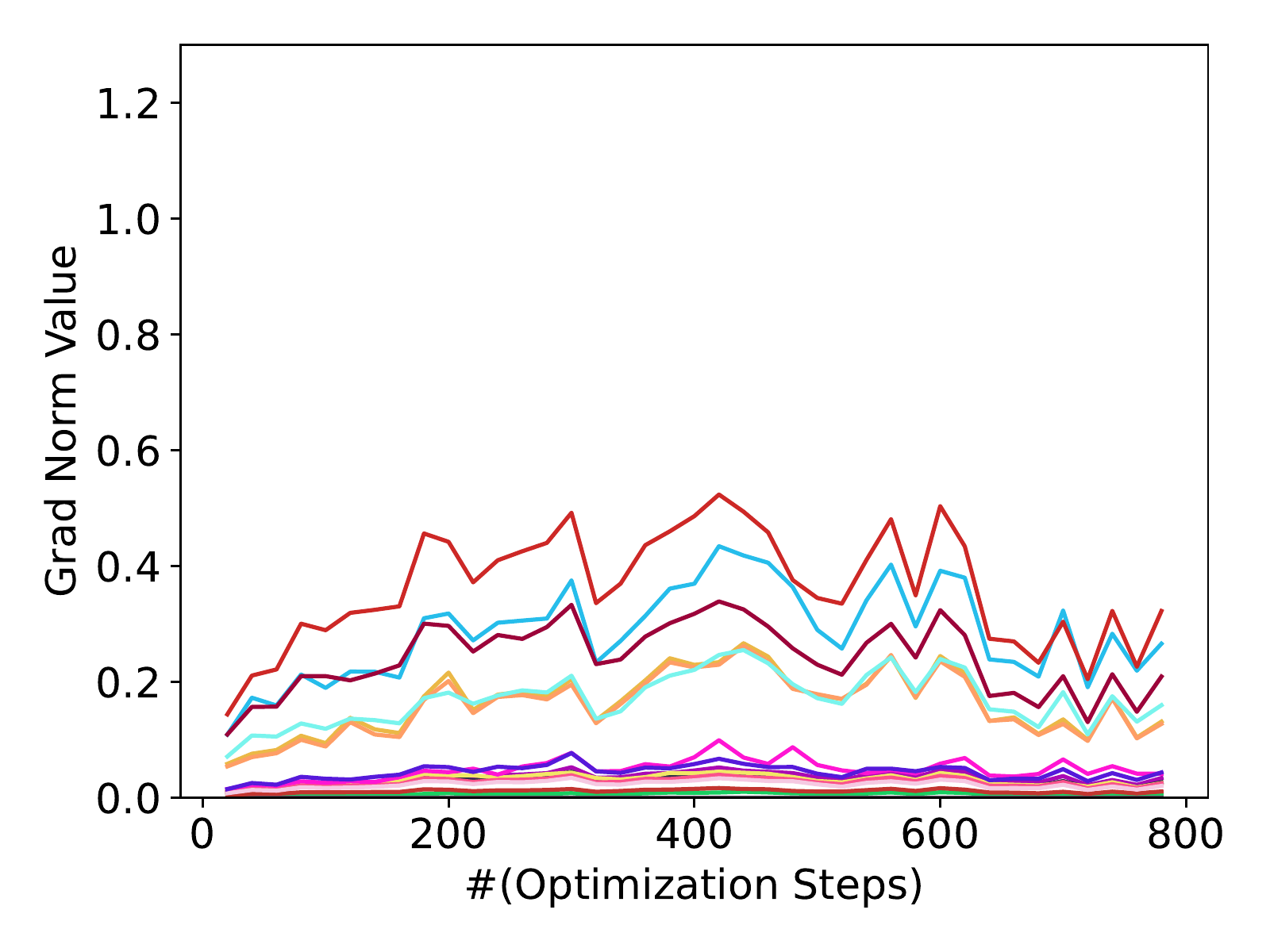}
\caption{{Success Run} }
\label{fig: success_run_grad_norm}
\end{subfigure}

\caption{Gradient norm across different parameters at layer 22 on RTE dataset fine-tuning.  Here the left figure (``Failed Run'') refers to the case when the fine-tuned model cannot beat the  majority classifier and the right figure (``Success Run'') otherwise. Different colors represent different parameters. Due to space limitation, legends are omitted and we cannot show results from all layers. But in our observation, most layers have similar phenomenon. }
\label{fig: gradient_norm_across_time}
\vspace{-3mm}
\end{figure*}

\vspace{-2mm}
\paragraph{Failure of Gradual Unfreezing} From \cref{fig: perf_over_time}, the accuracy of gradual unfreezing is significantly  worse than full fine-tuning,\footnote{As pointed out by \citet{mosbach2020stability}, without bias correction, the original results in \citet{devlin-etal-2019-bert} paper can be pretty bad and unstable, so we add bias correction on top of \citet{devlin-etal-2019-bert} to make a strong baseline.} although it indeed achieves smaller update to pretrained model parameters compared with full fine-tuning (\cref{fig: raw_norm_over_time}).

\vspace{-2mm}
\paragraph{Convergence Racing between Parameters} To investigate the reason behind this unsatisfying performance of GU, we plot the component-wise maximum update (measured by component-wise maximum rooted mean squared difference,\footnote{We also plot the parameter change measured by cosine distance and put the figures in \cref{app: cosine_measurement}} as different layers can have multiple components with different dimensions) at each layer in \cref{fig: inc_norm_over_time}. 
Clearly, from the very beginning, the parameter updates for both early-tuned parameters and newly-unfrozen  parameters have quickly diminished in GU so not many updates happen later when new parameters join. 
Based on this observation, we hypothesize that the failure of GU is because the early-tuned parameters have already converged in early iterations and cannot be re-activated later to adapt for newly-unfrozen  parameters. 

To verify this hypothesis, we simply modify GU to stop using early-tuned parameters in the previous iteration and instead copy the weight from the pretrained BERT model. 
For simplicity, we term this method as ``GU (restart)''. 
Note that full fine-tuning is just the last iteration of GU (restart) when all layers are unfrozen. 
We plot GU (restart) performance in \cref{fig: perf_over_time}, and observe that GU (restart) achieves consistently much better performance than GU.
However, GU (restart) is much more unstable than GU, and the best performance is only reached when almost all layers have been tuned.

\vspace{-2mm}
\paragraph{Unbalanced Gradients across Parameters} 
To investigate the cause of the convergence racing problem, we analyze the distribution of parameters' gradients from different components, by plotting the gradient norm in 
\cref{fig: gradient_norm_across_time}.\footnote{In \cref{fig: success_run_grad_norm} and \cref{fig: failed_run_grad_norm}, we do the fine-tuning by following the recommended setting in ~\cite{devlin-etal-2019-bert}.} Here we follow the terminology in previous work \cite{dodge2020fine, mosbach2020stability} that, if a fine-tuned model checkpoint can't beat the majority classifier,\footnote{Here ``majority classifier'' means simply using the majority labels in the training dataset as the predicted label.} then it is a ``Failed Run'' and ``Success Run'' otherwise.  Comparing \cref{fig: success_run_grad_norm} and \cref{fig: failed_run_grad_norm}, we can find that convergence racing still exists under normal full fine-tuning. The success run wins by making all parameters update roughly following the same trends and making gradient norms well-bounded.  

\vspace{-2mm}
\paragraph{Discussion on Fine-Tuning Stability} 
From the case study over GU, we observe that there is a convergence racing problem between parameters of model components, which would lead to incompetent or unstable performance. 
Based on existing studies, we argue that a robust PLM fine-tuning method should satisfy the following principles: 
\begin{enumerate}
  \item Updating the full set of parameters, not just a subset of layers. 
  \item No significant parameter updates to avoid \textit{catastrophic forgetting} problem. 
  \item Adjusting the convergence speed for parameters to mitigate the \textit{convergence racing}. 
\end{enumerate}
\citet{mosbach2020stability} proposed to use small learning rate (e.g., $1e^{-5}$) to mitigate \textit{catastrophic forgetting}, together with longer fine-tuning process.
However, unbalanced gradients across different components is another factor for \textit{catastrophic forgetting} and \textit{convergence racing}, which is not resolved by only using small learning rates.

\begin{table*}[ht!]
\centering
\begin{tabular}{@{}cccccccccc@{}}
\toprule
\multirow{2}{*}{Approach} & \multicolumn{3}{c}{RTE} & \multicolumn{3}{c}{MRPC} & \multicolumn{3}{c}{COLA} \\ \cmidrule(l){2-10} 
 & Std & Mean & Max & Std & Mean & Max & Std & Mean & Max \\ \cmidrule(l){1-10}
\citet{devlin-etal-2019-bert} & 4.5 & 50.9 & 67.5 & 3.9 & 84.0 & 91.2 & 25.6 & 45.6 & 64.6 \\
{ + bias correction} & 4.0 & 70.6 & 75.5 & 2.1 & 89.2 & \textbf{92.2} & 11.1 & 59.2 & 64.1 \\
\citet{lee2019mixout} & 7.9 & 65.3 & 74.4 & 3.8 & 87.8 & 91.8 & 20.9 & 51.9 & 64.0 \\
\citet{mosbach2020stability} & 2.7 & 67.3 & 71.1 & 0.8 & 90.3 & {91.7} & 1.8 & 62.1 & \textbf{65.3} \\
CWGNC & \textbf{1.3} & \textbf{73.1} & \textbf{75.1} & \textbf{0.6} & \textbf{90.5} & {91.5} & \textbf{1.7} & \textbf{62.2} & {65.0} \\ \bottomrule
\end{tabular}%
\caption{Experiment results for fine-tuning benchmark. Boldfaced numbers are best under each criterion.}
\label{tab:grad_clip_results}
\vspace{-2mm}
\end{table*}

\section{The Component-Wise Gradient Norm Clipping Method}
\label{sec:method}

Bearing these principles in mind, we propose the component-wise gradient norm clipping (CWGNC) method. 
The components are different parameters serving different functionalities. 
For example, in a Transformer-based neural architecture, components can be the weight matrices of Key-Query-Value in Transformer architectures, the weight and bias terms in feed-forward layers, etc. 
For those optimizers which maintain first-order and/or second-order bias correction terms (e.g., Adam \cite{kingma2015adam} and AdamW \cite{loshchilov2018decoupled} as the popular optimizer in PLM literature), our proposed norm clipping operation happens before the bias correction terms computation and does not interfere with normal bias correction process.

By clipping the gradient norm of each component individually, we aim to balance the distribution of gradients across different components to adjust their convergence speed, hence mitigating the convergence racing problem.

\section{Experiments}
\label{sec:experiment}
To evaluate our CWGNC method, 
we follow \citet{mosbach2020stability} to run our CWGNC method over $25$ different runs and report aggregated results on the validation set of three different datasets: RTE (Acc), MRPC (F1) and COLA (MCC). We report the standard deviation (Std), averaged performance (Mean) and maximum performance (Max).  Hyperparameters are tuned on held-out data. More implementation details are in \cref{app: implement}.

\paragraph{Baseline Setting} For baselines, we first consider the original BERT paper reported results  \cite{devlin-etal-2019-bert} and we run original BERT fine-tuning with bias correction to make a stronger baseline following the observation in \cite{mosbach2020stability}. We also compare to Mixout regularization methods \cite{lee2019mixout} and ``simple but hard-to-beat baseline'' proposed by \citet{mosbach2020stability}. 

\paragraph{Experiment Results}
Experiment results are shown in \cref{tab:grad_clip_results}. Here we can see our CWGNC achieves significantly better performance in terms of averaged performance and standard deviation. Compared with previous state-of-the-art results from \citep{mosbach2020stability}, our method only needs to tune $5$ epochs and does not need to wait for $20$ epochs even on these small datasets so our method indeed converges faster. 

Our method is also robust to a wide range of the selection of gradient norm clipping thresholds. We show this in \cref{tab:threshold_sensitivity} on COLA dataset as we find that fine-tuning methods are particularly unstable on COLA dataset. Here we see that within a reasonable range of threshold ($<1$), the model performance would be mostly maintained and the standard deviation is well controlled. If we use a significantly larger threshold ($>=1$), less control is enforced by CWGNC and it will degrade to normal full model fine-tuning if we further increase the threshold. 

\begin{table}[htbp!]
\small
\centering
\begin{tabular}{@{}cccc@{}}
\toprule
\multirow{2}{*}{Threshold} & \multicolumn{3}{c}{COLA} \\ \cmidrule(l){2-4} 
 & Std & Mean & Max  \\ \cmidrule(l){1-4}
0.01 & 1.3 & 61.6 & 64.3  \\
0.05 & 1.7 & 62.2 & 65.0  \\
0.1 & 1.4 & 61.4 & 65.1  \\
0.5 & 1.3 & 61.6 & 64.1 \\
1 & {1.3} & {61.4} & {63.8}  \\
5 & 15.4 & 57.3 & 65.0 \\ \bottomrule
\end{tabular}%
\caption{CWGNC fine-tuning results on COLA under different gradient norm clipping thresholds.}
\label{tab:threshold_sensitivity}
\vspace{-3mm}
\end{table}

\section{Conclusion}
In this paper, we investigate the instability problem of fine-tuning large pretrained language models. Inspired by previous works, we first experiment with gradual unfreezing methods, which should help minimize the updates in top layers and ease the catastrophic forgetting problem. However, further experiment results do not support that and we find it is because there is a convergence racing issue between different parameters, namely that early-converged parameters can limit the search space for other parameters. Based on this finding, we propose to do component-wise gradient norm clipping and achieve significant improvement on averaged performance, smaller standard deviation and quicker convergence speed. Our method is robust to the selection of gradient norm clipping threshold. In the future, we will try to study whether the component racing problem also exists in pretrained language models with different sizes and different pretraining methods. 

\section{Limitations}
In this paper we mainly work with one particularly popular large pretrained language model BERT~\cite{devlin-etal-2019-bert}. While we believe our empirical investigation and conclusion is widely applicable to a wide range of current transformer-based large pretrained language models, more experiments and theoretical explanations are needed for further research. Also, due to computational resources limitation, we cannot investigate whether the most recent large models like T0 \cite{sanh2021multitask} are stable under fine-tuning. We follow the evaluation protocols in previous works \cite{dodge2020fine, lee2019mixout, mosbach2020stability} to investigate the instability of fine-tuning on small datasets including COLA, RTE and MRPC. But for real challenging low-resource situation, we believe it can be more complicated and more investigation is needed. 

\section*{Acknowledgements}
This material is based on research sponsored by Air Force Research Laboratory (AFRL) under agreement number FA8750-19-1-1000. The U.S. Government is authorized to reproduce and distribute reprints for Government purposes notwithstanding any copyright notation therein.
The views and conclusions contained herein are those of the authors and should not be interpreted as necessarily representing the official policies or endorsements, either expressed or implied, of Air Force Laboratory, DARPA or the U.S. Government.

\bibliography{emnlp2022}

\begin{thebibliography}{10}
\expandafter\ifx\csname natexlab\endcsname\relax\def\natexlab#1{#1}\fi

\bibitem[{Devlin et~al.(2019)Devlin, Chang, Lee, and
  Toutanova}]{devlin-etal-2019-bert}
Jacob Devlin, Ming-Wei Chang, Kenton Lee, and Kristina Toutanova. 2019.
\newblock \href {https://doi.org/10.18653/v1/N19-1423} {{BERT}: Pre-training of
  deep bidirectional transformers for language understanding}.
\newblock In \emph{Proceedings of the 2019 Conference of the North {A}merican
  Chapter of the Association for Computational Linguistics: Human Language
  Technologies, Volume 1 (Long and Short Papers)}, pages 4171--4186,
  Minneapolis, Minnesota. Association for Computational Linguistics.

\bibitem[{Dodge et~al.(2020)Dodge, Ilharco, Schwartz, Farhadi, Hajishirzi, and
  Smith}]{dodge2020fine}
Jesse Dodge, Gabriel Ilharco, Roy Schwartz, Ali Farhadi, Hannaneh Hajishirzi,
  and Noah Smith. 2020.
\newblock Fine-tuning pretrained language models: Weight initializations, data
  orders, and early stopping.
\newblock \emph{arXiv preprint arXiv:2002.06305}.

\bibitem[{Howard and Ruder(2018)}]{howard2018universal}
Jeremy Howard and Sebastian Ruder. 2018.
\newblock Universal language model fine-tuning for text classification.
\newblock In \emph{Proceedings of the 56th Annual Meeting of the Association
  for Computational Linguistics (Volume 1: Long Papers)}, pages 328--339.

\bibitem[{Kingma and Ba(2015)}]{kingma2015adam}
Diederik~P Kingma and Jimmy Ba. 2015.
\newblock Adam: A method for stochastic optimization.
\newblock In \emph{ICLR (Poster)}.

\bibitem[{Kirkpatrick et~al.(2017)Kirkpatrick, Pascanu, Rabinowitz, Veness,
  Desjardins, Rusu, Milan, Quan, Ramalho, Grabska-Barwinska
  et~al.}]{kirkpatrick2017overcoming}
James Kirkpatrick, Razvan Pascanu, Neil Rabinowitz, Joel Veness, Guillaume
  Desjardins, Andrei~A Rusu, Kieran Milan, John Quan, Tiago Ramalho, Agnieszka
  Grabska-Barwinska, et~al. 2017.
\newblock Overcoming catastrophic forgetting in neural networks.
\newblock \emph{Proceedings of the national academy of sciences},
  114(13):3521--3526.

\bibitem[{Lee et~al.(2019)Lee, Cho, and Kang}]{lee2019mixout}
Cheolhyoung Lee, Kyunghyun Cho, and Wanmo Kang. 2019.
\newblock Mixout: Effective regularization to finetune large-scale pretrained
  language models.
\newblock In \emph{International Conference on Learning Representations}.

\bibitem[{Loshchilov and Hutter(2018)}]{loshchilov2018decoupled}
Ilya Loshchilov and Frank Hutter. 2018.
\newblock Decoupled weight decay regularization.
\newblock In \emph{International Conference on Learning Representations}.

\bibitem[{Mosbach et~al.(2020)Mosbach, Andriushchenko, and
  Klakow}]{mosbach2020stability}
Marius Mosbach, Maksym Andriushchenko, and Dietrich Klakow. 2020.
\newblock On the stability of fine-tuning bert: Misconceptions, explanations,
  and strong baselines.
\newblock In \emph{Proceedings of ICLR}.

\bibitem[{Sanh et~al.(2022)Sanh, Webson, Raffel, Bach, Sutawika, Alyafeai,
  Chaffin, Stiegler, Scao, Raja, Dey, Bari, Xu, Thakker, Sharma, Szczechla,
  Kim, Chhablani, Nayak, Datta, Chang, Jiang, Wang, Manica, Shen, Yong, Pandey,
  Bawden, Wang, Neeraj, Rozen, Sharma, Santilli, Fevry, Fries, Teehan,
  Biderman, Gao, Bers, Wolf, and Rush}]{sanh2021multitask}
Victor Sanh, Albert Webson, Colin Raffel, Stephen~H. Bach, Lintang Sutawika,
  Zaid Alyafeai, Antoine Chaffin, Arnaud Stiegler, Teven~Le Scao, Arun Raja,
  Manan Dey, M~Saiful Bari, Canwen Xu, Urmish Thakker, Shanya~Sharma Sharma,
  Eliza Szczechla, Taewoon Kim, Gunjan Chhablani, Nihal Nayak, Debajyoti Datta,
  Jonathan Chang, Mike Tian-Jian Jiang, Han Wang, Matteo Manica, Sheng Shen,
  Zheng~Xin Yong, Harshit Pandey, Rachel Bawden, Thomas Wang, Trishala Neeraj,
  Jos Rozen, Abheesht Sharma, Andrea Santilli, Thibault Fevry, Jason~Alan
  Fries, Ryan Teehan, Stella Biderman, Leo Gao, Tali Bers, Thomas Wolf, and
  Alexander~M. Rush. 2022.
\newblock Multitask prompted training enables zero-shot task generalization.
\newblock \emph{Proceedings of ICLR}.

\bibitem[{Wolf et~al.(2020)Wolf, Debut, Sanh, Chaumond, Delangue, Moi, Cistac,
  Rault, Louf, Funtowicz, Davison, Shleifer, von Platen, Ma, Jernite, Plu, Xu,
  Scao, Gugger, Drame, Lhoest, and Rush}]{wolf-etal-2020-transformers}
Thomas Wolf, Lysandre Debut, Victor Sanh, Julien Chaumond, Clement Delangue,
  Anthony Moi, Pierric Cistac, Tim Rault, Rémi Louf, Morgan Funtowicz, Joe
  Davison, Sam Shleifer, Patrick von Platen, Clara Ma, Yacine Jernite, Julien
  Plu, Canwen Xu, Teven~Le Scao, Sylvain Gugger, Mariama Drame, Quentin Lhoest,
  and Alexander~M. Rush. 2020.
\newblock \href {https://www.aclweb.org/anthology/2020.emnlp-demos.6}
  {Transformers: State-of-the-art natural language processing}.
\newblock In \emph{Proceedings of the 2020 Conference on Empirical Methods in
  Natural Language Processing: System Demonstrations}, pages 38--45, Online.
  Association for Computational Linguistics.

\end{thebibliography}
\bibliographystyle{acl_natbib}
\clearpage
\appendix

\section{Gradual Unfreezing Algorithm}
\label{sec: algo_GU}
The gradual unfreezing algorithm we implemented following \citet{howard2018universal} is shown in \cref{alg:gradual_unfreezing}.

\begin{algorithm}
\caption{Gradual Unfreezing Fine-Tuning}\label{alg:gradual_unfreezing}
\begin{algorithmic}
\Require {A multi-layer Transformer PLM $M$ with its $L$-layer parameters $\{\theta^{(i)}, 1\leq i \leq L\}$. The maximum number of iterations $T$, the dataset $D$, the optimizer ${Opt}$, the number of epochs for each iteration $E$}.
\Ensure $1 \leq T \leq L$
\State $\hat{R}^{(0)} \gets \phi$ 
\Comment{$\hat{R}^{(l)}$ represents to-be-tuned parameters at $l$-th iteration}.
\While{$T \leq L$}
\State $\hat{R}^{(T)} \gets \hat{R}^{(T-1)} \cup \{\theta^{(L-T+1)}\}$
\For{$j = 1 \to E$}
\State $\mathcal{L} \gets$ {Forward}$(M, D)$
\State $G \gets$ {Backward}$(\mathcal{L}, L-T+1, L)$
\State \Comment{Only needs to compute the gradient for top-$T$ layers}
\State {Update}$(Opt, G, \hat{R}^{(T)})$
\State \Comment{Apply $G$ over $\hat{R}^{(T)}$}
\State {Replace}$(M, \hat{R}^{(T)}, L-T+1, L)$  
\State \Comment{Replace the updated layers back to make sure the updated parameters are involved in next epoch forward process}
\EndFor
\State $T \gets T + 1$
\EndWhile
\end{algorithmic}
\end{algorithm}

\section{Implementation Details}
\label{app: implement}
Our codebase is based on Huggingface Transformers \cite{wolf-etal-2020-transformers} example fine-tuning scripts and will be released later. We tune models using our method for $5$ epochs. For weight decay and warm-up steps, we follow the settings original fine-tuning method as described in \citep{devlin-etal-2019-bert}. We here report the result with clipping threshold $0.05$ as we empirically find it works better on held-out dataset. We also show later that our method is actually pretty robust to a wide range of threshold picking in \cref{tab:threshold_sensitivity}.

\section{Gradient Update Measured by Cosine Similarities}
\label{app: cosine_measurement}
\begin{figure}[htbp!]

\centering
\includegraphics[width=\columnwidth]{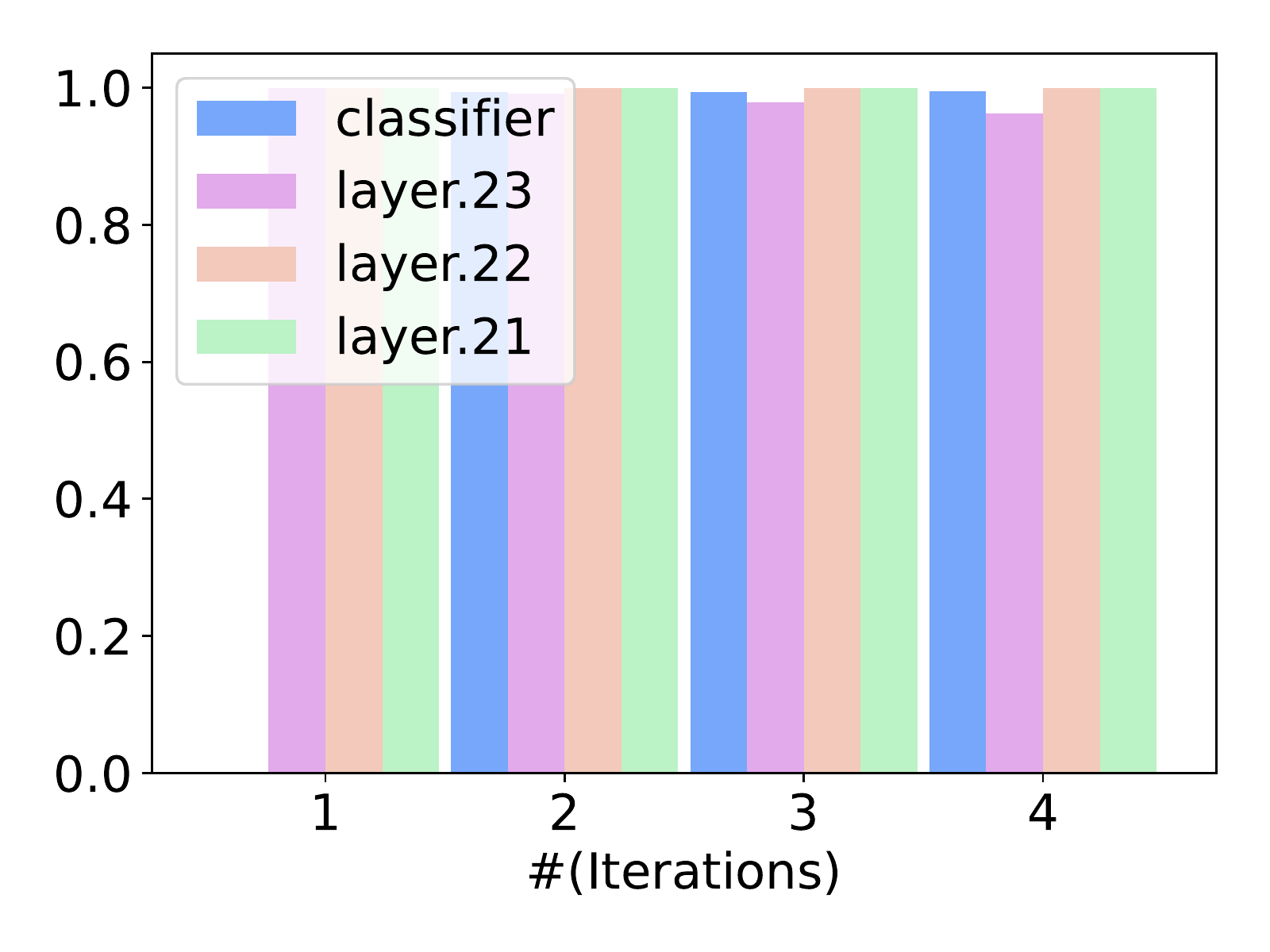}
\caption{The scales of \textbf{incremental} parameter update at each iteration in GU (\textbf{compared with previous iteration, measured by cosine similarity}). As different layers can have different components with different dimensionalities, we show the component-wise maximum rooted mean squared difference for each layer. We only plot the first $4$ iterations as we observe that the performance is almost stable by the  $4$-th iteration. $22$-th layer (``layer.21'' in the figure) update is too small to plot out in this figure.}
\label{fig: cos_inc_norm_over_time}
\end{figure}

\begin{figure}[htbp!]

\centering
\includegraphics[width=\columnwidth]{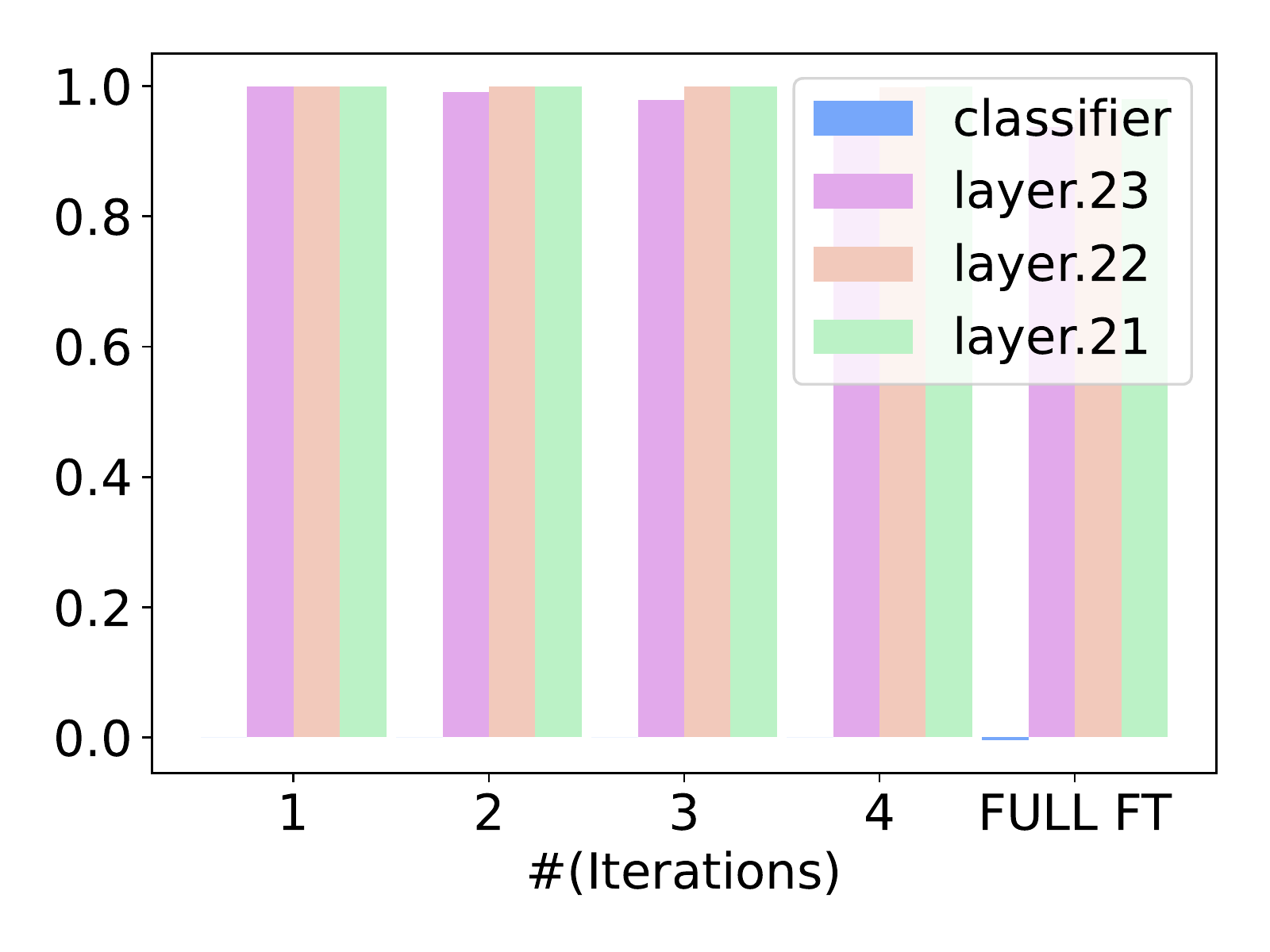}
\caption{The scales of parameter update at each iteration in GU (\textbf{compared with original BERT model, measured by cosine similarity}). Too small updates cannot be seen in this figure. We only plot the first $4$ iterations as we observe that the performance is almost stable in $4$-th iteration.}
\label{fig: cos_raw_norm_over_time}
\end{figure}

In the main text we measures the update via root-mean-square difference. Here we show that if measured by cosine-similarity, we would obtain similar conclusion. Here, \cref{fig: cos_inc_norm_over_time} is the cosine similarity version for \cref{fig: inc_norm_over_time}. \cref{fig: cos_raw_norm_over_time} is the cosine similarity version for \cref{fig: raw_norm_over_time}. Note that because smaller cosine similarity indicates more changes, in contrast to square-mean-root difference, we use the component-wise  \textbf{minimum} cosine similarity to represent the update at each layer.


\end{document}